\documentclass[runningheads]{llncs}

\usepackage{eccv}

\usepackage{eccvabbrv}

\usepackage{graphicx}
\usepackage{booktabs}

\usepackage[accsupp]{axessibility}  
\usepackage{multirow}
\usepackage{algorithm}
\usepackage{algpseudocode}  
\algrenewcommand\algorithmicrequire{\textbf{Input:}}
\algrenewcommand\algorithmicensure{\textbf{Output:}}

\usepackage[pagebackref,breaklinks,colorlinks,citecolor=eccvblue]{hyperref}

\usepackage{orcidlink}

\begin{document}

\title{Leveraging Phase Information to Boost Unrolled Network Learning for Image Deblurring} 

\titlerunning{UPADNet}

\author{Samira Malek\inst{1}\orcidlink{0009-0005-2530-3846} \and
Haichuan Zhang\inst{1}\orcidlink{0009-0003-1678-6415} \and
Chul Lee\inst{2}\orcidlink{0000-0001-9329-7365} \and
Vishal Monga\inst{1}\orcidlink{0000-0002-5100-2263}}

\authorrunning{S.~Malek et al.}

\institute{Pennsylvania State University, University Park, PA 16802, USA
\email{\{sxm6547,hzz5333,vum4\}@psu.edu}\\
\and
Dongguk University, Seoul 04620, South Korea\\
\email{chullee@dongguk.edu}}
\maketitle
\begin{abstract}
While most image deblurring techniques directly restore the spatial image variable, we propose an amplitude and phase decomposition recognizing the importance of accurate phase estimation in recovering sharp image details. To that end, we first develop novel linear minimum mean squared (LMMSE) estimators of the amplitude and phase of the blurred, noisy image observation. An iterative optimization algorithm follows that recovers the sharp image using the aforementioned LMMSE estimators. Finally, matrix parameters that are statistically determined and fixed in the iterative algorithm are now learned using a training dataset of clean and degraded observations. Our deblurring engine is dubbed UPADNet -- Unrolled Phase and Amplitude Decomposition Network, such that each iteration of the underlying phase and amplitude recovery algorithm is parameterized and trained end-to-end. Experiments over benchmark evaluation datasets such as GoPro, RealBlur and  COCO datasets confirm that UPADNet outperforms state of the art deep networks including those based on algorithm unrolling in the image domain. The benefits of UPADNet are even more pronounced in high noise and limited training data regimes.     
\end{abstract}
\section{Introduction}
\label{sec:intro}
Image deblurring, a sub-problem of image restoration, aims to recover a sharp image from a blurred observation.  In real-world imaging, blur artifacts arise from various sources such as atmospheric turbulence, diffraction, optical defocusing, and camera motion\cite{kundur1996blind}.

Various approaches have been developed to address blind image deblurring, one of which is leveraging mathematical modeling and statistical estimation techniques \cite{kundur1996blind}. Some methods focus on estimating the blur kernel using sharp edge prediction \cite{joshi2008psf} and sparsity constraints in the image domain \cite{xu2013unnatural, pan2016l_0}. A common strategy has been to incorporate strong regularizations, such as $\ell_0$-norm sparsity \cite{xu2013unnatural, pan2016l_0}, normalized sparsity \cite{krishnan2011blind}, and total variation constraints \cite{perrone2015clearer}, to mitigate the ill-posed nature of the problem. Several studies have explored novel priors, including patch priors \cite{sun2013edge}, dark channel priors \cite{pan2017deblurring}, and extreme channel priors \cite{yan2017image}, which exploit natural image statistics to enhance sharp image estimation. 
Bayesian inference has also been widely used, employing marginalization over the high-dimensional image space \cite{wipf2014revisiting} and general sparse image priors \cite{babacan2012bayesian} to improve robustness. A notable contribution in this line is \cite{shan2008high}, which introduced a unified probabilistic framework for joint blur kernel estimation and image restoration. Framelet-based methods \cite{cai2011framelet} offer alternative strategies by leveraging transform-domain representations. Some works analyze the fundamental limitations of existing marginal likelihood optimization approaches and provide a broader theoretical evaluation of blind deconvolution algorithms, introducing efficient MAP-based methods \cite{levin2011understanding, levin2011efficient}.
In~\cite{huang2021effective}, an iterative refinement framework was proposed that categorizes kernel types by alternating between kernel and image updates to achieve improved results. A modified Newton integration algorithm is presented from a control perspective, offering noise tolerance and fast convergence \cite{liao2021modified}.
Additionally, in~\cite{delbracio2021polyblur}, Polyblur was introduced as a non-iterative method for removing mild blur by leveraging polynomial reblurring, which offers practical benefits for lightly degraded images. 

Recent advances in image deblurring have largely focused on data-driven approaches tailored to specific datasets. Deep neural networks have shown remarkable success in this domain \cite{sarker2021deep,lecun2015deep}. Convolutional neural networks such as \cite{nah2017deep, tao2018scale, cho2021rethinking} adopt multi-scale architectures with stacked sub-networks, progressively enhancing sharpness in a coarse-to-fine manner. To mitigate the computational burden of multi-scale upsampling and the diminishing returns of deeper fine-scale networks, the Deep Multi-Patch Hierarchical Network (DMPHN) was proposed in \cite{zhang2019deep}. Similarly, MPRNet \cite{zamir2021multi} leverages multi-stage hierarchical refinement, while MAXIM \cite{tu2022maxim} introduces an MLP-based backbone for deblurring. NAFNet \cite{chen2022simple} further demonstrates that strong performance can be achieved even without nonlinear activations. Transformer-based architectures, including Restormer \cite{zamir2022restormer}, Uformer \cite{wang2022uformer}, and Stripformer \cite{tsai2022stripformer}, improve performance by modeling long-range dependencies via attention mechanisms. Frequency-domain approaches such as MRLPFNet \cite{dong2023multi}, the efficient frequency transformer \cite{kong2023efficient}, and frequency selection analysis in \cite{mao2023intriguing} aim to better integrate spectral and spatial information for improved high-frequency detail preservation.

In parallel, the use of generative and adversarial models represents another line of approach. DeblurGAN \cite{kupyn2018deblurgan} employs an end-to-end framework based on a conditional GAN with content loss, while CGAN \cite{lin2019tell} adopts region focused strategies. Stochastic refinement-based approaches have also emerged, {notably probabilistic} deblurring \cite{whang2022deblurring}. Additionally, recent work has explored joint tasks of depth estimation and restoration from defocused images using deep networks, as demonstrated {in} \cite{nazir2023depth}, pixel-wise kernel prediction for blind motion deblurring~\cite{carbajal2023blind}, and enhanced defocus blur estimation using multi-interactive strategies \cite{li2024multi}. AIFNet \cite{ruan2021aifnet} contributes to this direction by using refocusing techniques and light field synthetic aperture. CodEx \cite{majee2022codex} features a modular framework that jointly addresses temporal deblurring and tomographic reconstruction.
These developments underline a growing trend toward hybrid models that integrate architectural innovations, spectral priors, and task specific insights to enhance restoration fidelity, robustness, and generalization. 

Despite these advancements, fully deep neural network-based methods often suffer from key limitations: they require large amounts of annotated data for effective training, have limited generalizability across diverse imaging conditions, and demand extensive computational resources and training time, which can hinder their deployment in real-world and resource-constrained applications. Recently, algorithm unrolling \cite{monga2021algorithm, csahin2025unlocking, shlezinger2023model} has gained attention as an effective strategy for addressing image deconvolution. This approach combines the interpretability of traditional iterative optimization with the adaptability of deep learning by unfolding an iterative algorithm—such as Half-Quadratic Splitting~(HQS) {\cite{li2020efficient,li2019algorithm}}—into a learnable network where key parameters, like filters or step sizes, are trained. Unrolling methods have been applied successfully to various deblurring contexts, including networks based on iterative variational Bayesian methods \cite{huang2022unrolled}, second-order (Newton) updates for image restoration \cite{10820096}, photon-limited scenarios \cite{sanghvi2022photon}, non-uniform blind deblurring \cite{richmond2022non}, and low-light image enhancement in joint frameworks \cite{vo2025deep}. Additionally, kernel-based techniques like the Gaussian Kernel Mixture Network enhance defocus deblurring by embedding blur priors into deep models \cite{quan2021gaussian}.

Even with significant advances in deep and unrolled networks for image deblurring, the problem remains fundamentally ill-posed, making reliable high-quality restoration under diverse real-world degradations challenging. Accurate phase recovery is particularly critical, as phase encodes the structural and geometric information of an image \cite{oppenheim1981importance,lim1990two}. Prior work \cite{8954150} has demonstrated the importance of phase-only representations for kernel initialization in image deblurring. Motivated by these insights, we propose a principled phase–amplitude decomposition framework that explicitly models and estimates both components in the Fourier domain. We first derive novel \textbf{linear minimum mean squared error (LMMSE) estimators} for the amplitude and phase of blurred, noisy observations. An iterative optimization algorithm is then developed to recover the latent sharp image using these estimators. Unlike classical formulations where statistical parameters are fixed, we learn the underlying matrices directly from paired clean and degraded data. This leads to \textbf{UPADNet} (Unrolled Phase and Amplitude Decomposition Network), where each iteration of the phase–amplitude recovery algorithm is parameterized and trained end-to-end. Extensive experiments on GoPro and RealBlur demonstrate that UPADNet \textbf{outperforms state-of-the-art} deep and unrolled methods. Moreover, our model exhibits superior \textbf{robustness under high noise}, camera motion, and atmospheric degradations, and maintains strong performance in low-data training regimes, highlighting its generalization capability. The code is available at \href{https://github.com/SamiraMalek/Phase-Unrolling}{GitHub}.

\textbf{Notation:} Unless stated otherwise, uppercase letters (\eg,~$A$) represent matrices throughout the paper. The symbol $*$ denotes circular convolution, {and} $\circ$ {is} the Hadamard (element-wise) product. 
The element in the $i$-th row and $j$-th column of a matrix ${A}$ is denoted by $A_{ij}$. 
The notation $\frac{{A}}{{B}}$ denotes element-wise division between matrices ${A}$ and ${B}$, that is,
$
\left(\frac{{A}}{{B}}\right)_{ij} = \frac{A_{ij}}{B_{ij}},
$
for all valid indices $(i,j)$. Let the Fourier transform of \({x}\) be denoted by \(\mathcal{F}\{{x}\} = X = A_X e^{j\theta_X}\), where \(A_X\) and \(\theta_X\) are the amplitude and phase of \(X\), respectively.
\section{LMMSE Estimation of Blurred Phase and Amplitude}
\label{sec:Ph-Amp Decomposition}
Image deblurring is classically modeled as a linear shift-invariant convolution process corrupted by additive noise:
\begin{equation}
\label{eq:debluring-model}
    {z} = {h} * {u} + {n}
\end{equation}
where ${u}$ is the latent sharp image, ${h}$ the blur kernel, and ${n}$ additive noise. This forward model is standard in computational imaging and inverse problems \cite{kundur1996blind,levin2009understanding,chan2005fundamentals}.
In the Fourier domain, convolution becomes element-wise multiplication:
\begin{equation}
\label{eq:Four-debluring-model}
   Z = H \circ U + N 
\end{equation}
Let $Z = A_Z e^{j\theta_Z}$ denote the polar form of the Fourier transform. It has long been known that the phase encodes structural information while the amplitude encodes energy distribution \cite{oppenheim1981importance,lim1990two}. 

A naive yet commonly used estimation of the Fourier-domain amplitude and phase of the observed blurry image $Z$ is obtained directly from the convolution property in the frequency domain. Specifically, let $A_H$, $\theta_H$ and $A_U$, $\theta_U$ denote the amplitudes and phases of the blur kernel $H$ and the latent sharp image $U$, respectively. Then, the naive estimates of the amplitude $\tilde{A}_Z$ and phase $\tilde{\theta}_Z$ of $Z$ are given by:
\begin{equation}
\tilde{A}_Z = A_H \circ A_U,
\quad
\tilde{\theta}_Z = \theta_H + \theta_U.
\end{equation}
In noiseless convolution case, \ie, $N=0$, we have $A_Z = \tilde{A}_Z$ and $\theta_Z = \tilde{\theta}_Z$. 
However, in realistic settings with noise and kernel uncertainty,  these equalities no longer hold. Further, in recent investigations, the image blur is known to depart from the standard convolution model \cite{nah2017deep,rim2020real}.
To address these challenges and motivated by linear minimum mean squared error (LMMSE) estimation theory \cite{kay1993fundamentals}, we aim to estimate \(A_Z\) and \(\theta_Z\) via generalized linear operators: 
\begin{equation}
\label{eq:estimators}
\hat{A}_Z = \mathcal{W}_1 \circ A_H \circ A_U + \mathcal{W}_2, \quad \hat{\theta}_Z = \mathcal{W}_3 \circ \theta_H + \mathcal{W}_4 \circ \theta_U + \mathcal{W}_5
\end{equation}
such that these linear estimations minimize the squared errors:
\begin{equation}
   E[\|A_Z - \hat{A}_Z\|_2^2] 
   \quad \text{and} \quad 
   E[\|\theta_Z - \hat{\theta}_Z\|_2^2].
\end{equation}
The following theorems characterize the optimal element-wise linear estimators.
    \begin{theorem}[Amplitude Estimation]
    \label{thrm:amp}
        Let ${A}_Z$ and $ {A}_H \circ {A}_U$ be random matrices with finite means and variances. 
        \begin{equation}
 \min_{{\mathcal{W}}_1, {\mathcal{W}}_2} E[\|{A}_Z - \hat{{A}}_Z\|_F^2],\quad s.t.  \; \hat{{A}}_Z = {\mathcal{W}}_1 \circ {A}_H \circ {A}_U + {\mathcal{W}}_2  
\end{equation}
is minimized when:
\begin{equation}
    [{\mathcal{W}}_1^*]_{ij} = \frac{\mathrm{Cov}([{A}_Z]_{ij}, [{A}_H]_{ij} [{A}_U]_{ij})}{\mathrm{Var}([{A}_H]_{ij} [{A}_U]_{ij})}, 
\end{equation}
\begin{equation}
   [{\mathcal{W}}_2^*]_{ij} = \mathbb{E}[[{A}_Z]_{ij}] - [{\mathcal{W}}_1^*]_{ij} \mathbb{E}[[{A}_H]_{ij}  [{A}_U]_{ij}] 
\end{equation}
\textbf{Proof:} See supplementary material Section 1. 
    \end{theorem}
\begin{theorem}[Phase Estimation]
\label{thrm:phase}
Let ${\theta}_Z$, ${\theta}_H$, and ${\theta}_U$ be random matrices with finite means and variances. Consider:
\begin{equation}
\min_{{\mathcal{W}}_3, {\mathcal{W}}_4, {\mathcal{W}}_5} 
\mathbb{E}\!\left[\|{\theta}_Z - \hat{{\theta}}_Z\|_F^2\right], \quad \text{s.t. } 
\hat{{\theta}}_Z 
= {\mathcal{W}}_3 \circ {\theta}_H 
+ {\mathcal{W}}_4 \circ {\theta}_U 
+ {\mathcal{W}}_5 .
\end{equation}
The minimizer satisfies, for every entry $(i,j)$:
\begin{align}
[{\mathcal{W}}_3^*]_{i,j}
&= \frac{
\sigma^{2}_{U,i,j}\, \gamma_{H,i,j} 
- \sigma_{HU,i,j}\, \gamma_{U,i,j}
}{\Delta_{i,j}}, \\
[\,{\mathcal{W}}_4^*\,]_{i,j}
&= \frac{
- \sigma_{HU,i,j}\, \gamma_{H,i,j} 
+ \sigma^{2}_{H,i,j}\, \gamma_{U,i,j}
}{\Delta_{i,j}}, \\
[\,{\mathcal{W}}_5^*\,]_{i,j}
&= \mu_{Z,i,j}
- [{\mathcal{W}}_3^*]_{i,j}\, \mu_{H,i,j}
- [{\mathcal{W}}_4^*]_{i,j}\, \mu_{U,i,j},
\end{align}
where:
\begin{align}
\sigma^{2}_{H,i,j} &= \operatorname{Var}([{\theta}_H]_{i,j}), &
\sigma^{2}_{U,i,j} &= \operatorname{Var}([{\theta}_U]_{i,j}), \nonumber\\
\sigma_{HU,i,j} &= \operatorname{Cov}([{\theta}_H]_{i,j},[{\theta}_U]_{i,j}), &
\gamma_{H,i,j} &= \operatorname{Cov}([{\theta}_Z]_{i,j},[{\theta}_H]_{i,j}), \nonumber\\
\gamma_{U,i,j} &= \operatorname{Cov}([{\theta}_Z]_{i,j},[{\theta}_U]_{i,j}),&
\Delta_{i,j} &= \sigma^{2}_{H,i,j}\sigma^{2}_{U,i,j} - \sigma_{HU,i,j}^2, \nonumber\\
\mu_{Z,i,j} &= \mathbb{E}[\,[{\theta}_Z]_{i,j}], &
\mu_{H,i,j} &= \mathbb{E}[\,[{\theta}_H]_{i,j}], \nonumber\\
\mu_{U,i,j} &= \mathbb{E}[\,[{\theta}_U]_{i,j}].\nonumber
\end{align}
\textbf{Proof:} See supplementary material Section 2. 
\end{theorem}

\begin{figure}[tb]
  \centering
  \begin{subfigure}{0.15\linewidth}
    \centering
    \includegraphics[width=\linewidth]{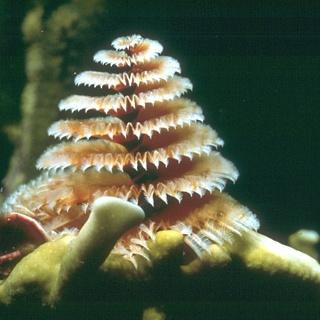}
    \caption{GT}
    \label{fig:GT-decomposition}
  \end{subfigure}
\hfill
  \begin{subfigure}{0.15\linewidth}
    \centering
    \includegraphics[width=\linewidth]{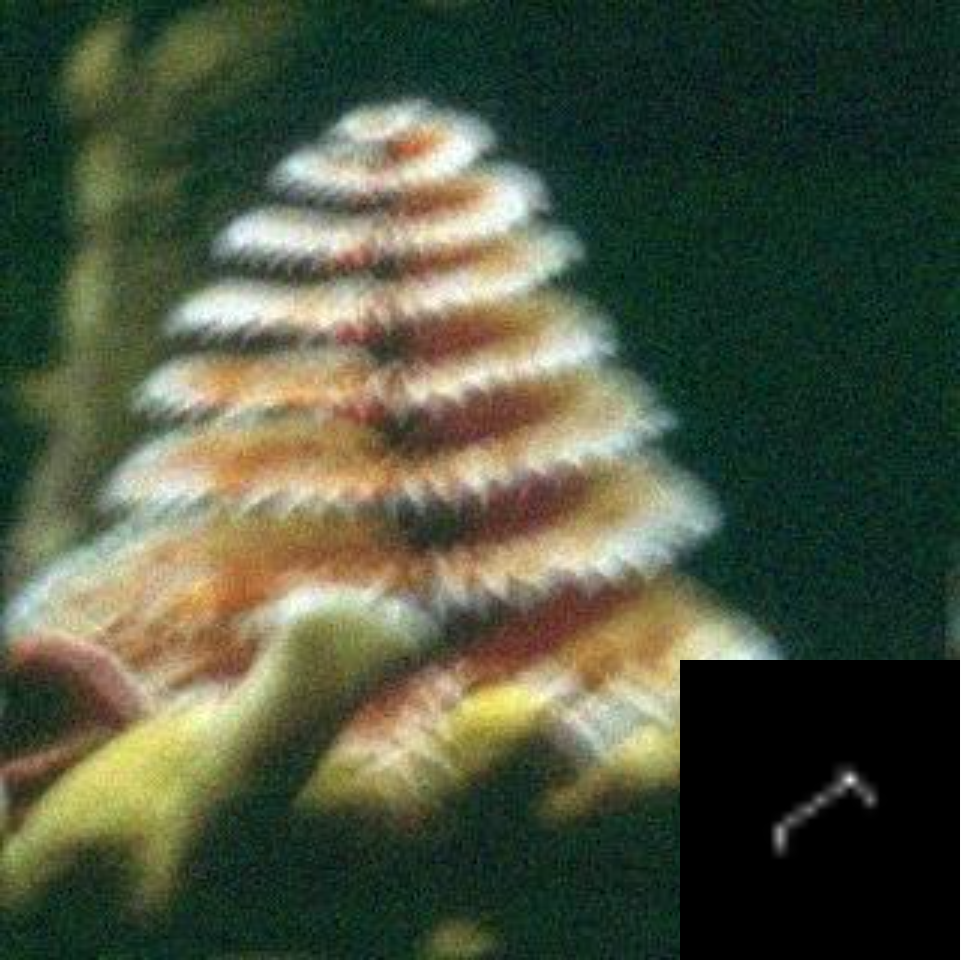}
    \caption{Blurred}
    \label{fig:blurred-decomposition}
  \end{subfigure} 
  \hfill
  \begin{subfigure}{0.33\linewidth}
    \centering
    \includegraphics[width=\linewidth]{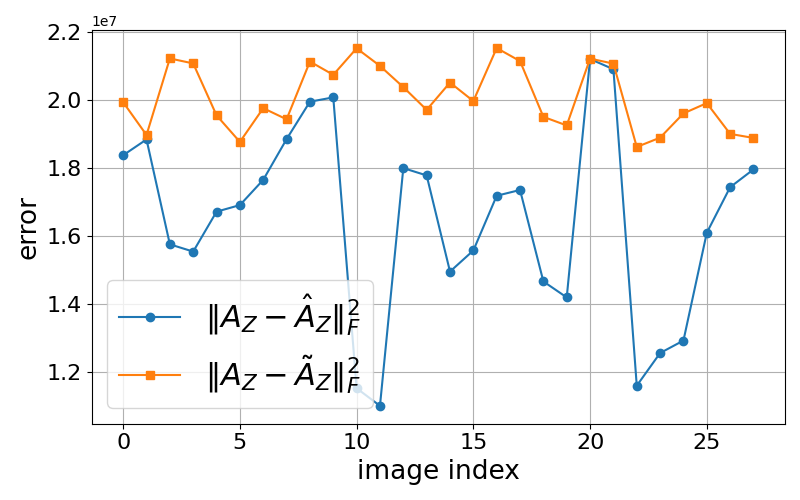}
    \caption{Amplitude error}
    \label{fig:amp-error}
  \end{subfigure}
  \hfill
  \begin{subfigure}{0.33\linewidth}
    \centering
    \includegraphics[width=\linewidth]{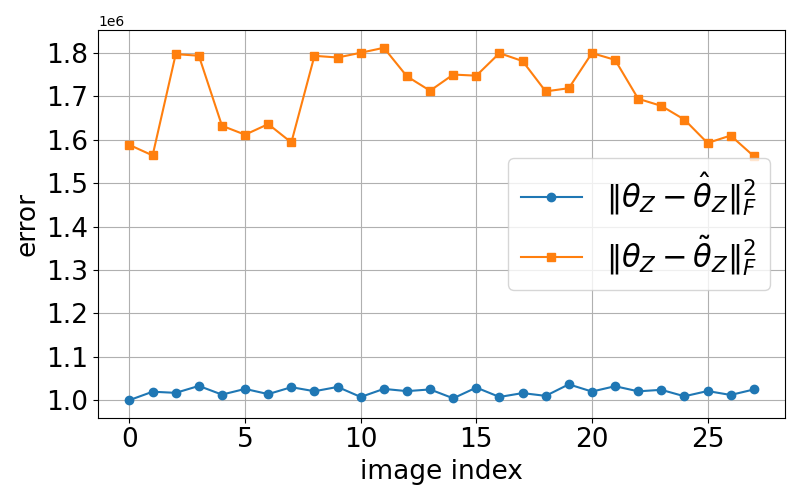}
    \caption{Phase error}
    \label{fig:ph-error}
  \end{subfigure} 
  \caption{Estimation of amplitude and phase components of the blurred observation. 
(a) Ground-truth sharp image. 
(b) Blurred observation generated by convolving the sharp image with a known kernel (bottom right of the blurred image) and addiing noise. 
(c) and (d) Comparison of the estimation errors of the naive and proposed LMMSE estimators for amplitude and phase on the test set. The proposed LMMSE estimators consistently produce lower estimation errors.}
  \label{fig:ph-amp-decomposition}
\end{figure}

Based on the aforementioned theorems, the proposed estimators achieve a lower expected MSE than the naive estimators, since the latter are a special case of the former. This result is further validated through experiments. Specifically, we evaluate the naive and proposed LMMSE estimators and verify that:
\begin{equation}
 E[\|A_Z - \tilde{A}_Z\|_2^2] \ge E[\|A_Z - \hat{A}_Z\|_2^2], \quad E[\|\theta_Z - \tilde{\theta}_Z\|_2^2] \ge E[\|\theta_Z - \hat{\theta}_Z\|_2^2]   
\end{equation}
To this end, we generate 600 blurred images by convolving 200 BSDS images~\cite{arbelaez2010contour} with three kernels from~\cite{li2020efficient} and adding Gaussian noise ($\sigma = 0.05$) to empirically estimate the statistical parameters $\{\mathcal{W}_i\}_{i=0}^5$ in \eqref{eq:estimators}. For evaluation, we use 28 images (14 images with two unseen kernels and additive Gaussian noise  with $\sigma = 0.05$. One example of the ground-truth image, blur kernel, and the resulting blurred image is shown in \cref{fig:GT-decomposition,fig:blurred-decomposition}. 
\cref{fig:amp-error,fig:ph-error} show the amplitude and phase errors, respectively. The proposed LMMSE estimators consistently achieve lower error than the naive multiplicative/additive models. This experimentally confirms that although convolution is multiplicative in the Fourier domain, the induced amplitude and phase statistics under noise deviate from the naive relations, and the proposed estimators significantly improves estimation of amplitude and phase. 

\section{Image Deblurring Via Phase and Amplitude Optimization}
\label{sec:iter-alg}
We estimate the latent sharp image by minimizing a regularized objective that enforces statistical consistency with the linear phase and amplitude predictors derived from \cref{eq:estimators} in Sec.~\ref{sec:Ph-Amp Decomposition}:
\begin{align}
\label{eq:objective}
\min_{A_U, A_H, \theta_U, \theta_H} \quad &
\underbrace{\left\| A_Z - (\mathcal{W}_1 \circ A_H \circ A_U  + \mathcal{W}_2) \right\|_F^2
+ \alpha \|A_H\|_F^2 + \beta \|M \circ A_U\|_{1,1}}_{\textbf{Amplitude loss}} \nonumber \\
& + \underbrace{\left\| \theta_Z - (\mathcal{W}_3 \circ \theta_H + \mathcal{W}_4 \circ \theta_U + \mathcal{W}_5) \right\|_F^2
+ \gamma \|\theta_H\|_F^2 + \mu \|\theta_U\|_{1,1}}_{\textbf{Phase loss}}
\end{align}
The first two terms of amplitude and phase losses enforce amplitude and phase data fidelity under the statistically optimal linear model, mitigating the errors introduced by noise and kernel uncertainty in naive Fourier inversion \cite{levin2009understanding}. Solving this objective yields $(A_U,\theta_U)$ from which the restored image is obtained via the inverse Fourier transform. The $\ell_1$ regularization on $M \circ A_U$ promotes sparsity in the Fourier-domain representation of natural images, with the mask $M$ acting as a frequency-domain filter \cite{rudin1992nonlinear,hyvarinen2009natural}. In contrast, the blur kernel is expected to be smooth; therefore, an $\ell_2$ penalty is imposed on $\theta_H$ and $A_H$, consistent with classical regularization for ill-posed inverse problems \cite{tikhonov1977solutions,bertero1998introduction}. Moreover, incorporating an $\ell_1$ regularization term on $\theta_U$ enforces structural symmetry in the optimization updates, which is crucial for the unrolled architecture introduced in \cref{sec:unrolled-alg}. This mixed $\ell_1$--$\ell_2$ formulation reflects the structural asymmetry between images (sparse and edge-dominated) and kernels (smooth and compact), stabilizing blind deblurring while preserving high-frequency detail.

\begin{sloppypar}
The objective in \cref{eq:objective} is nonconvex. Following classical blind deconvolution strategies \cite{kundur1996blind,levin2009understanding}, we adopt an alternating optimization scheme that updates one variable at a time while fixing the others. This decomposes the problem into a sequence of simpler subproblems that admit closed-form or proximal solutions. Moreover, the $\ell_1$ regularization terms are non-smooth and therefore non-differentiable, which prevents direct closed-form minimization of \cref{eq:objective}. To address this, we employ HQS \cite{german1995nonlinear}, introducing auxiliary variables $S$ and $P$ to decouple the non-smooth penalties from the quadratic data-fidelity terms. This leads to six subproblems per iteration:
\begin{align}
A_H^{(t+1)} &= 
\arg\min_{A_H}
\|A_Z - \mathcal{W}_1 \circ A_H \circ A_U^{(t)} - \mathcal{W}_2\|_2^2 
+ \alpha \|A_H\|_2^2, \\
S^{(t+1)} &= 
\arg\min_{S}
\zeta \| M \circ S \|_1 
+ \beta\| A_U^{(t)} - S\|_2^2, \\
A_U^{(t+1)} &= 
\arg\min_{A_U}
\|A_Z - \mathcal{W}_1 \circ A_H^{(t+1)} \circ A_U - \mathcal{W}_2\|_2^2 
+ \beta \| A_U - S^{(t+1)}\|_2^2, \\
\theta_H^{(t+1)} &= 
\arg\min_{\theta_H}
\|\theta_Z - \mathcal{W}_3 \circ \theta_H - \mathcal{W}_4 \circ \theta_U^{(t)} - \mathcal{W}_5\|_2^2 
+ \gamma \|\theta_H\|_2^2, \\
P^{(t+1)} &= 
\arg\min_{P}
\xi \|P\|_1 
+ {\mu}\|\theta_U^{(t)} - P\|_2^2, \\
\theta_U^{(t+1)} &= 
\arg\min_{\theta_U}
\|\theta_Z - \mathcal{W}_3 \circ \theta_H^{(t+1)} - \mathcal{W}_4 \circ \theta_U - \mathcal{W}_5\|_2^2 
+ \mu \|\theta_U - P^{(t+1)}\|_2^2.
\end{align}
Each quadratic subproblem admits a closed-form solution, whereas the $\ell_1$-regularized auxiliary updates correspond to soft-thresholding proximal operators.\footnote{
The soft-thresholding operator is defined element-wise as 
$\operatorname{soft}(x,\lambda) = \mathrm{sign}(x)\max(|x|-\lambda,0)$.} 
The resulting closed-form update rules are summarized in Algorithm~\ref{alg:simplified-opad}. 
\end{sloppypar}

\begin{algorithm}[t]
\caption{Phase and Amplitude Decomposition Image Deblurring Algorithm}
\label{alg:simplified-opad}
\begin{algorithmic}[1]
\Require $A_Z, \theta_Z, \{\mathcal{W}_i\}_{i=1}^5$, regularization weights $\alpha, \beta, \gamma, \mu, \xi, \zeta$
\State \textbf{Initialize:} $A_U^{(0)}, A_H^{(0)}, \theta_U^{(0)}, \theta_H^{(0)}$
\For{$t = 0$ to $T$}

    \Statex \textbf{Amplitude updates:}
    \State \hspace{1em}$
    A_H^{(t+1)} =
    \frac{
      \left( \mathcal{W}_1 \circ A_U^{(t)} \right)\circ\left( A_Z -  \mathcal{W}_2 \right)
    }{
      \left( \mathcal{W}_1 \circ A_U^{(t)}  \right)^2 + \alpha
    }$
    \State \hspace{1em}$
    S^{(t+1)} = \operatorname{soft}\!\left( A_U^{(t)},\frac{\zeta \circ |M|}{2\beta } \right)$
    \State \hspace{1em}$
    A_U^{(t+1)} =
    \frac{
      \left( \mathcal{W}_1 \circ A_H^{(t+1)}  \right)\circ
      \left( A_Z -  \mathcal{W}_2  + \beta S^{(t+1)} \right)
    }{
      \left( \mathcal{W}_1 \circ A_H^{(t+1)}  \right)^2 + \beta
    }$

    \Statex \textbf{Phase updates:}
    \State \hspace{1em}$
    \theta_H^{(t+1)} =
    \frac{
      \mathcal{W}_3 \circ
      \left( \theta_Z - \left( \mathcal{W}_4 \circ \theta_U^{(t)} + \mathcal{W}_5 \right) \right)
    }{
      \left(  \mathcal{W}_3 \right)^2 + \gamma
    }$
    \State \hspace{1em}$
    P^{(t+1)} =
    \operatorname{soft}\!\left(
      \theta_U^{(t)} , \frac{\xi}{2\mu }
    \right)$
    \State \hspace{1em}$
    \theta_U^{(t+1)} =
    \frac{
      \mathcal{W}_4 \circ
      \left( \theta_Z - \left( \mathcal{W}_3 \circ \theta_H^{(t+1)} + \mathcal{W}_5 \right) + \mu P^{(t+1)} \right)
    }{
      \left(  \mathcal{W}_4 \right)^2 + \mu
    }$

\EndFor
\Ensure $A_U^{(T+1)}, A_H^{(T+1)}, \theta_U^{(T+1)}, \theta_H^{(T+1)}$
\end{algorithmic}
\end{algorithm}

\section{Unrolled Phase and Amplitude Decomposition Image Deblurring Network (UPADNet)}
\label{sec:unrolled-alg}
\begin{figure}[tb]
  \centering

  \begin{subfigure}{0.88\linewidth}
    \centering
    \includegraphics[width=\linewidth]{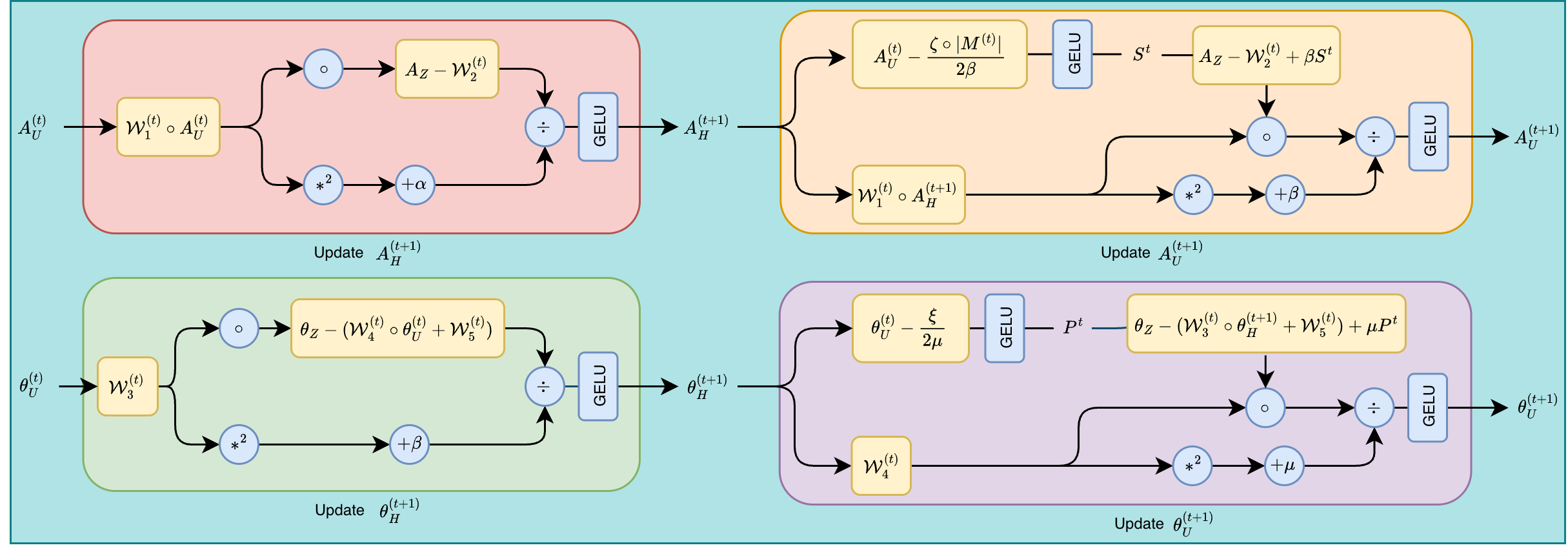}
    \caption{}
    \label{fig:UPADBlock}
  \end{subfigure}
  \hfill
  \begin{subfigure}{0.085\linewidth}
    \centering
    \includegraphics[width=\linewidth]{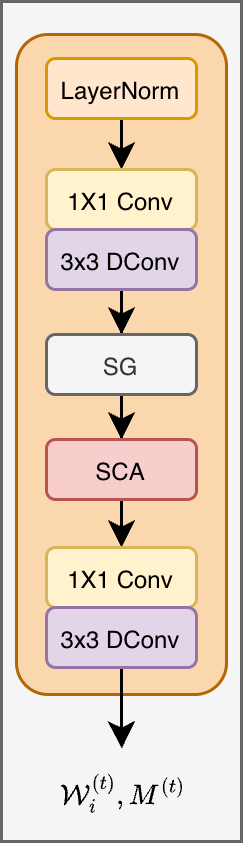}
    \caption{}
    \label{fig:Weight-Generator}
  \end{subfigure}
  \vspace{0.5em}
  \begin{subfigure}{0.99\linewidth}
    \centering
    \includegraphics[width=\linewidth]{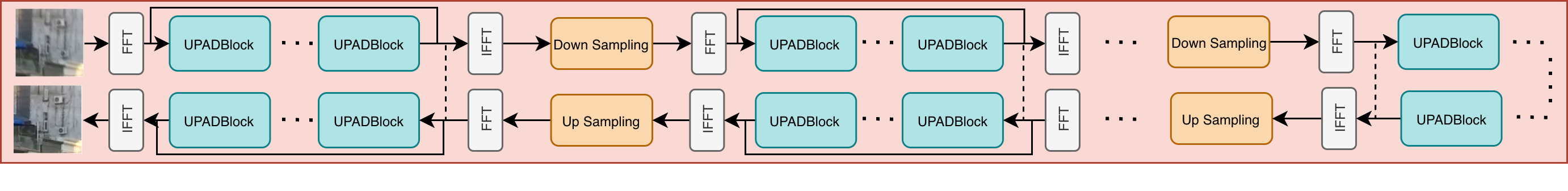}
    \caption{}
    \label{fig:UPADNet}
  \end{subfigure}
  \caption{
Overview of the proposed UPADNet architecture. 
(a) {UPADBlock}: one unrolled iteration derived from \cref{alg:simplified-opad}, consisting of four sub-blocks that update $A_U$, $A_H$, $\theta_U$, and $\theta_H$ with learnable parameters. 
(b) {Weight Generator}: dynamically produces iteration-specific weights $\{\mathcal{W}_i^t\}_{i=1}^5$ and filter $M^t$. 
(c) {UPADNet}: the complete multi-scale network obtained by stacking multiple UPADBlocks with intermediate downsampling and upsampling for blind deblurring.}
  \label{fig:UPADNet-detail}
\end{figure}

Since the optimal linear estimators $\{\mathcal{W}_i\}_{i=1}^5$ depend on unknown image and blur statistics, they are not directly available in practical blind deblurring scenarios. Moreover, fixed analytical parameters may be suboptimal under complex real-world degradations. Therefore, we learn these quantities from data and unroll the optimization procedure into a trainable architecture to improve robustness and reconstruction performance.
Specifically, we propose the \emph{Unrolled Phase and Amplitude Decomposition Image Deblurring Network (UPADNet)}, a model-driven deep architecture derived from the iterative optimization procedure in \cref{alg:simplified-opad}. Each iteration of the underlying algorithm is unfolded into a learnable module, referred to as a {UPADBlock} (see \cref{fig:UPADBlock}), where the parameters $\{\mathcal{W}_i^t\}_{i=1}^5$ and filter $M^t$ are learned from data. This design preserves the interpretability of the original optimization scheme while enabling data-adaptive refinement. Each UPADBlock consists of four sub-blocks that sequentially update the amplitude and phase components of the latent image and blur kernel, \ie, $A_U$, $A_H$, $\theta_U$, and $\theta_H$. This decomposition explicitly leverages the complementary roles of amplitude (energy distribution) and phase (structural information) in the Fourier domain. 

Since $A_U$ and $A_H$ are non-negative and $\theta_U, \theta_H \in [0,2\pi]$, we enforce positivity and smoothness through a nonlinear activation function (GELU) at the end of each sub-block. Moreover, the update $S^{(t+1)}$ corresponds to a soft-thresholding operator, 
which can be simplified under the constraint $A_U \ge 0$:
\begin{align}
S^{(t+1)} 
= \operatorname{soft}\!\left(A_U^{(t)}, \frac{\zeta \circ |M^t|}{2\beta}\right)
&= \mathrm{sign}(A_U^{(t)}) 
   \max\!\left(|A_U^{(t)}| - \frac{\zeta \circ |M^t|}{2\beta}, 0 \right)\nonumber\\
   &= \operatorname{ReLU}\!\left( A_U^{(t)} - \frac{\zeta \circ |M^t|}{2\beta}\right)\nonumber\\
   \label{eq:appox_soft}
   & \approx \operatorname{GELU}\!\left( A_U^{(t)} - \frac{\zeta \circ |M^t|}{2\beta}\right).
\end{align}
In~\eqref{eq:appox_soft}, to improve gradient flow and training stability, we replace ReLU with GELU, which provides a smoother approximation to the proximal operator. Similarly, the phase update is implemented as:
\begin{equation}
P^{(t+1)} 
\approx \operatorname{GELU}\!\left(
\theta_U^{(t)} - \frac{\xi}{2\mu}
\right).
\end{equation}

To estimate the unknown parameters in each iteration and enhance flexibility across iterations, we introduce a \emph{Weight Generator} module (see \cref{fig:Weight-Generator}) that dynamically produces the weights $\{\mathcal{W}_i^t\}$ and filter $M^t$ at each stage. The module incorporates the SimpleGate (SG) and Simplified Channel Attention~(SCA) mechanisms proposed in \cite{chen2022simple}, enabling lightweight yet expressive parameter modulation. This design ensures compatibility between generated weights and the intermediate feature dimensions while maintaining computational efficiency.

As each UPADBlock represents one iteration of \cref{alg:simplified-opad}, stacking multiple UPADBlocks forms the complete UPADNet architecture. To further improve the representation power and enlarge the effective receptive field, we adopt a multi-scale strategy: intermediate features are progressively downsampled after several blocks, and corresponding upsampling layers are applied toward the output stage (see \cref{fig:UPADNet}). 

\section{Experiments}
\label{sec:Exp}
We evaluate \textbf{UPADNet} on three widely used image deblurring benchmarks. 
\textbf{GoPro}~\cite{nah2017deep} is a large-scale synthetic dataset generated via frame accumulation, which implicitly includes object motion blur and complex motion trajectories, where the effective blur kernel is difficult to approximate using a simple parametric model. 
\textbf{RealBlur-R} and \textbf{RealBlur-J}~\cite{rim2020real} are real-world blur datasets with carefully aligned ground-truth images, primarily capturing low-light camera shake scenarios. To further investigate the behavior of UPADNet under specific degradations, \ie, camera motion and mild atmospheric turbulence effects consistent with the convolutional model in~\cref{eq:debluring-model}, we additionally construct a synthetic benchmark using \textbf{COCO}~\cite{lin2014microsoft}. Specifically, sharp images are convolved with real-world motion kernels extracted from~\cite{li2020efficient} to simulate motion-induced blur, followed by the addition of Gaussian noise to model sensor corruption. In total, 6000 images are generated for training and 1500 images for testing. This setup allows us to systematically evaluate robustness under controlled camera motion and noise conditions and to compare performance against both unrolled optimization-based networks and purely deep learning-based architectures, particularly in low-training or noise-corrupted regimes.

UPADNet is trained end-to-end using the AdamW optimizer with standard data augmentation strategies, including random cropping, horizontal and vertical flipping, and rotation. Training is performed on $256 \times 256$ patches, while evaluation is conducted on full-resolution images. The training--test split for the GoPro and RealBlur datasets follow those used in prior works~\cite{11095124, mao2024adarevd}. Quantitative performance is reported using the widely adopted PSNR and SSIM metrics.

\begin{table}[!t]
\centering
\caption{Quantitative comparison on GoPro \cite{nah2017deep} and RealBlur \cite{rim2020real}.}
\label{tab:GoPro-RealBlur}
\setlength{\tabcolsep}{3pt}
\renewcommand{\arraystretch}{1.05}

\begin{tabular}{l cc cc cc  c}
\toprule 
\multirow{2}{*}[-3pt]{\textbf{Method}} & \multicolumn{2}{c}{GoPro} & \multicolumn{2}{c}{RealBlur-R}
& \multicolumn{2}{c}{RealBlur-J}  & \multirow{2}{*}[-3pt]{Params (M)}\\
\cmidrule(lr){2-3}\cmidrule(lr){4-5}\cmidrule(lr){6-7}
& PSNR$\uparrow$ & SSIM$\uparrow$ & PSNR$\uparrow$ & SSIM$\uparrow$
  & PSNR$\uparrow$ & SSIM$\uparrow$ &  \\
\midrule
DeblurGAN-v2 \cite{kupyn2019deblurgan}   & 29.55 & 0.934 & 36.44 & 0.9347 & 29.69 & 0.8703 & 60.9 \\
DUBLID \cite{li2020efficient} & 29.96 & 0.940 & 35.35 & 0.925 & 28.42 & 0.860 & N/A \\ 
DeepSN-Net \cite{10820096} & 32.83 & 0.960 &36.22 & 0.957 & 29.30 & 0.895 & N/A\\
Stripformer \cite{tsai2022stripformer}        & 33.08  & 0.962 & 39.84 & 0.9737 & 32.48 & 0.9290 & 20.0 \\
NAFNet\cite{chen2022simple}    & 33.69 & 0.967 & 35.84 & 0.952& 27.94& 0.854 & 67.8 \\
UFPNet \cite{fang2023self}      & 34.06 & 0.968 & 39.84 & 0.974 & 32.48 & 0.929 & 80.3 \\
FFTformer \cite{kong2023efficient} &34.21&0.968& 40.11 &0.9753 &32.62 &0.9326&16.6\\
AdaRevD \cite{mao2024adarevd}  & 34.60  & 0.972 & 36.53 & 0.957 & 30.12 & 0.894 & 220.0 \\
EGDeblurring \cite{11094688} & 34.62 & 0.973 & N/A  & N/A & N/A   & N/A & 80.3 \\
EVSSM \cite{11095124}   & 34.51 & 0.9713 & 41.27 & 0.977 & 34.34 & 0.945 & 17.1 \\
UPADNet     & \textbf{34.63} & \textbf{0.975} & \textbf{41.29} & \textbf{0.979} & \textbf{34.35} & \textbf{0.946} & 27.0 \\
\bottomrule
\end{tabular}
\end{table}

\begin{figure}[!t]
  \centering
  \includegraphics[width=\linewidth]{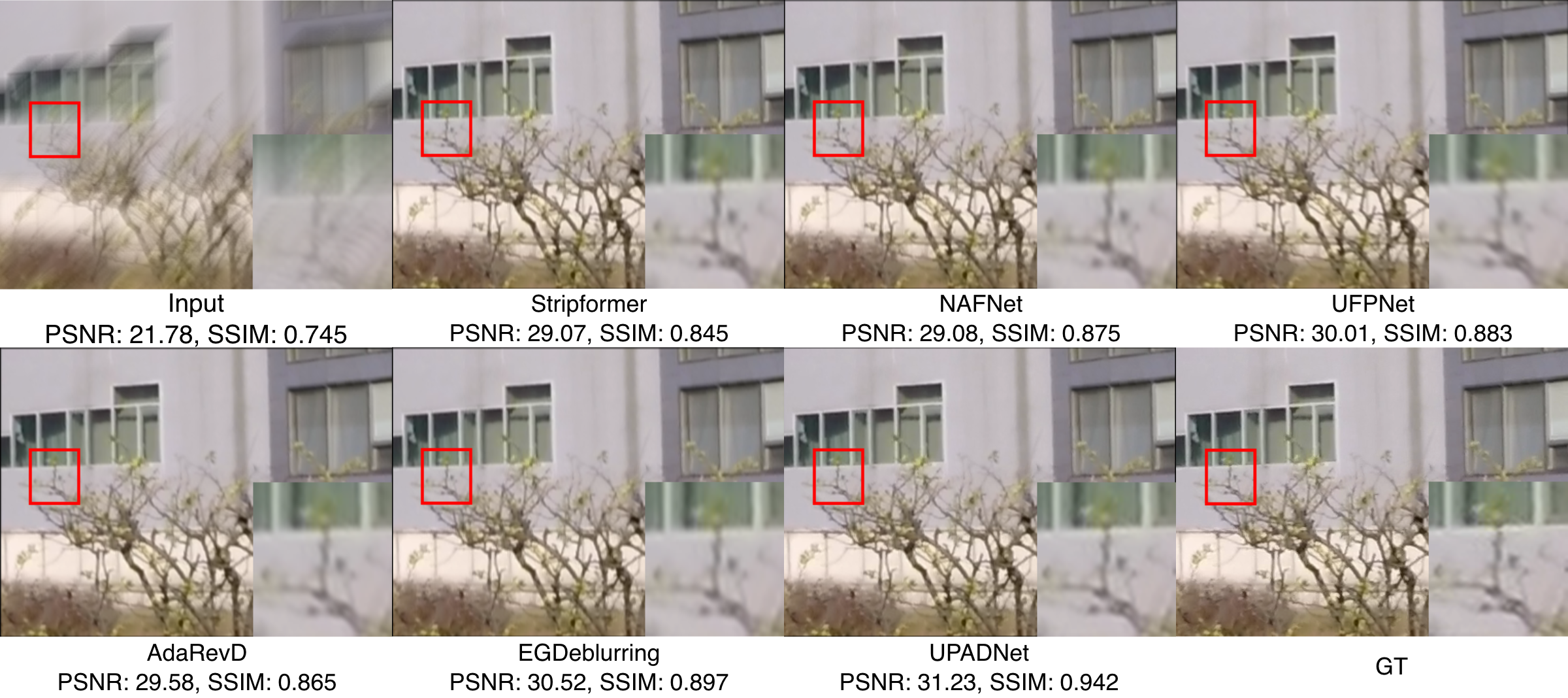}
  \caption{Qualitative results on the GoPro dataset.
  }
  \label{fig:Gopro1}
\end{figure}

Table~\ref{tab:GoPro-RealBlur} reports quantitative comparisons on the GoPro dataset against state-of-the-art deblurring methods. UPADNet achieves the best performance, reaching \textbf{34.63 dB} PSNR and \textbf{0.975} SSIM, outperforming recent transformer-based and diffusion-based approaches. Notably, our model requires only \textbf{27M} parameters, substantially fewer than large-scale models such as AdaRevD (220.8M) and HINet (88.7M), demonstrating a favorable accuracy–efficiency trade-off. Visual comparisons are provided in Fig.~\ref{fig:Gopro1}. UPADNet more effectively restores fine structures and high-frequency textures, particularly in regions containing sharp edges and repetitive patterns. The explicit phase–amplitude modeling enables the preservation of structural consistency while mitigating ringing artifacts. Table~\ref{tab:GoPro-RealBlur} also presents results on the RealBlur dataset. UPADNet achieves the best performance on both RealBlur-R and RealBlur-J, attaining \textbf{41.29/0.979} and \textbf{34.35/0.946} (PSNR/SSIM), respectively. The consistent improvements across both synthetic (GoPro) and real-world datasets demonstrate strong generalization capability. The performance gains on real data further suggest that the explicit Fourier-domain phase--amplitude decomposition enhances robustness to real camera shake and non-uniform blur, which are not perfectly captured by synthetic training data.

\begin{figure}[tb]
  \centering
  \includegraphics[width=\linewidth]{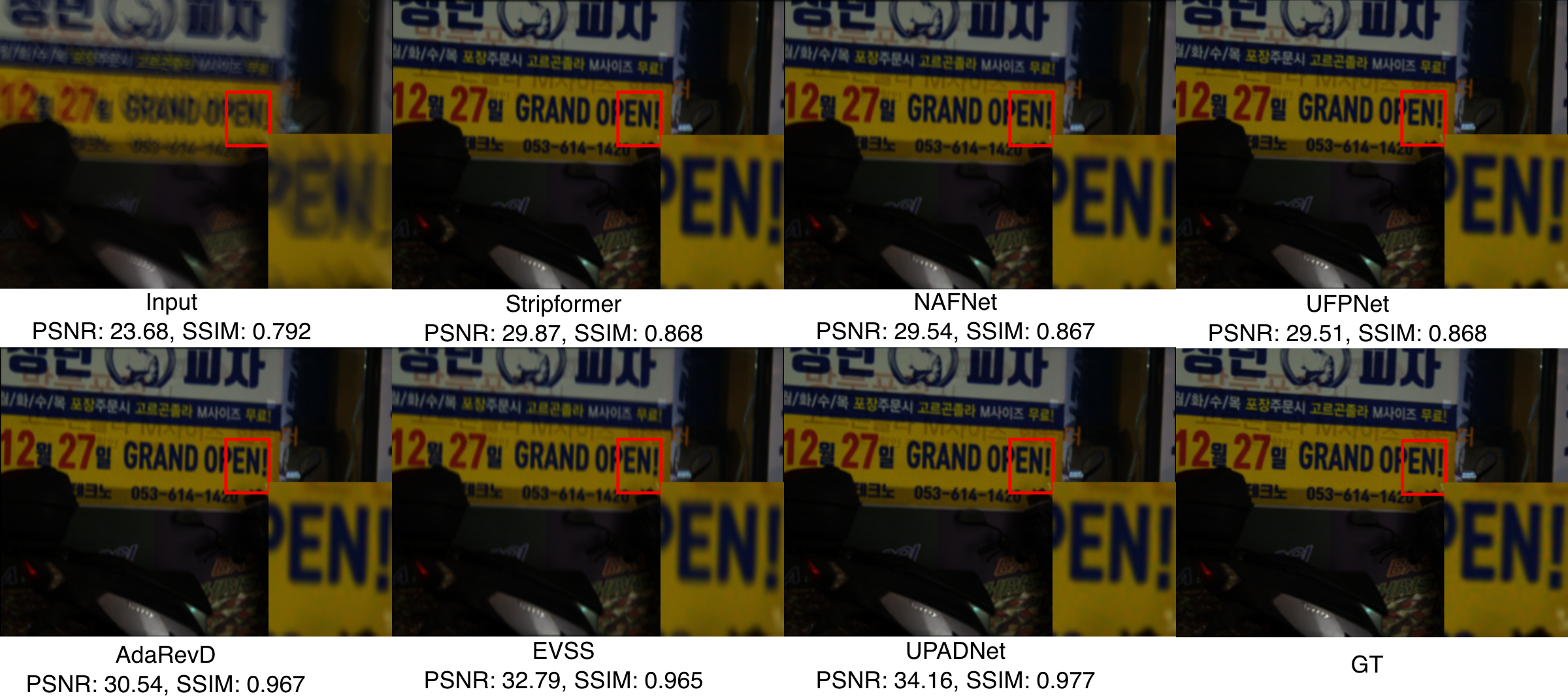}
  \caption{Qualitative results on the RealBlur-R dataset.
  }
  \label{fig:RealBlur-R}
\end{figure}

\begin{figure}[tb]
  \centering
  \includegraphics[width=\linewidth]{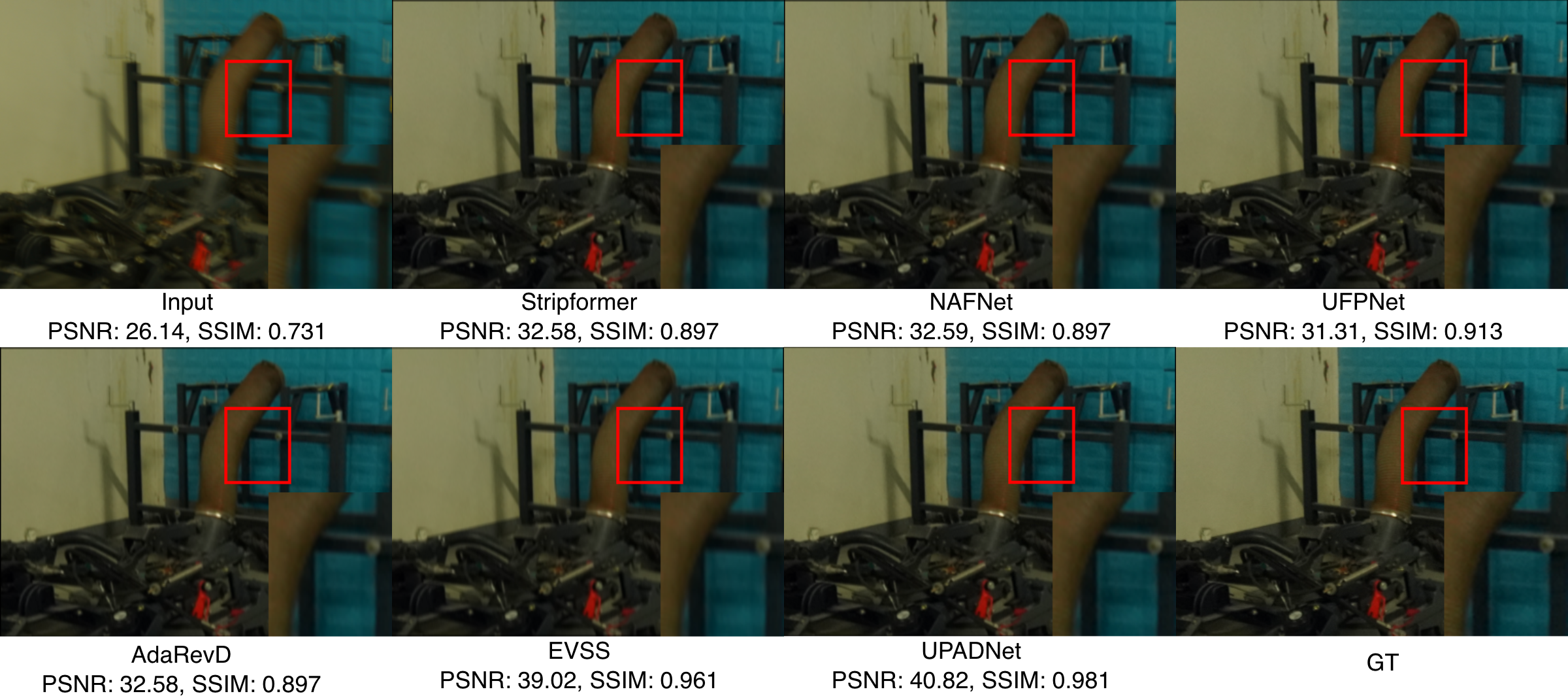}
  \caption{Qualitative results on the RealBlur-J dataset.
  }
  \label{fig:RealBlur-J}
\end{figure}


\begin{table}[t]
 \caption{Quantitative comparison with unrolled and deep network models under increasing levels of additive Gaussian noise on COCO.}
  \label{tab:noise_robustness}
\centering
\begin{tabular}{l cc cc cc }
\toprule 
\multirow{2}{*}[-3pt]{\textbf{Method}} & \multicolumn{2}{c}{$\sigma = 0.01$} & \multicolumn{2}{c}{$\sigma = 0.03$}
& \multicolumn{2}{c}{$\sigma = 0.05$}  \\
\cmidrule(lr){2-3}\cmidrule(lr){4-5}\cmidrule(lr){6-7}
& PSNR$\uparrow$ & SSIM$\uparrow$ & PSNR$\uparrow$ & SSIM$\uparrow$
  & PSNR$\uparrow$ & SSIM$\uparrow$ \\
\midrule
DUBLID \cite{li2020efficient} & 29.96 & 0.881 & 27.42 & 0.842 & 26.88 & 0.801  \\ 
Net-A\cite{richmond2022non}    & 27.20 & 0.835 & 26.63 & 0.798 & 25.94 & 0.756  \\
Stripformer \cite{tsai2022stripformer}        
& 31.84 & 0.921 
& 30.72 & 0.902 
& 29.95 & 0.874   \\
AdaRevD \cite{mao2024adarevd}  
& 32.10 & 0.925 
& 30.78 & 0.906 
& 30.11 & 0.879   \\
EVSSM \cite{11095124}   
& 31.62 & 0.918 
& 30.41 & 0.897 
& 30.01 & 0.872   \\
UPADNet     
& \textbf{32.58} & \textbf{0.932} 
& \textbf{31.47} & \textbf{0.914} 
& \textbf{30.86} & \textbf{0.891} \\
\bottomrule
\end{tabular}
\end{table}

\begin{figure}[!t]
  \centering
  \includegraphics[width=\linewidth]{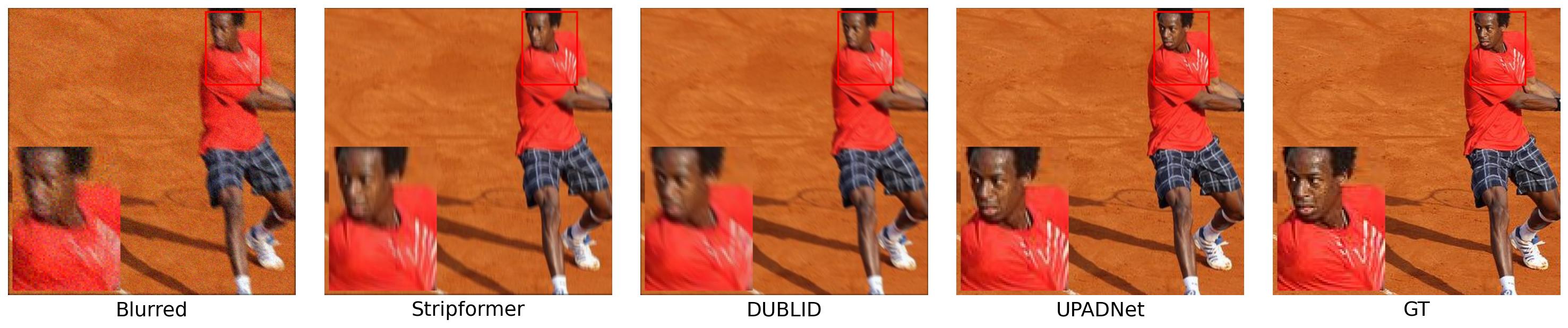}
  \caption{Deblurred results on COCO with $\sigma^2 = 0.05$.
  }
  \label{fig:COCO}
\end{figure}

\noindent \textbf{Noise Robustness (UPADNet vs. State of the Art):} To evaluate robustness under noisy conditions, we generate blurred images from COCO and add Gaussian noise with increasing variance ($\sigma^2 = 0.01, 0.03, 0.05$). Quantitative results are summarized in Table~\ref{tab:noise_robustness}. UPADNet consistently outperforms competing methods across all noise levels. While other state-of-the-art unrolled and deep models exhibit significant performance degradation as noise increases, UPADNet maintains stable performance, demonstrating strong noise robustness. This robustness can be attributed to the structured phase--amplitude decomposition and the linear estimation-inspired updates. Fig.~\ref{fig:COCO} further illustrates qualitative comparisons. UPADNet produces sharper reconstructions, especially in textured regions. Moreover, we compare UPADNet with Net-A and DUBLID, both of which are unrolled networks derived from similar optimization frameworks but without explicit phase--amplitude decomposition. 

\begin{figure}[tb]
  \centering
  \includegraphics[width=\linewidth]{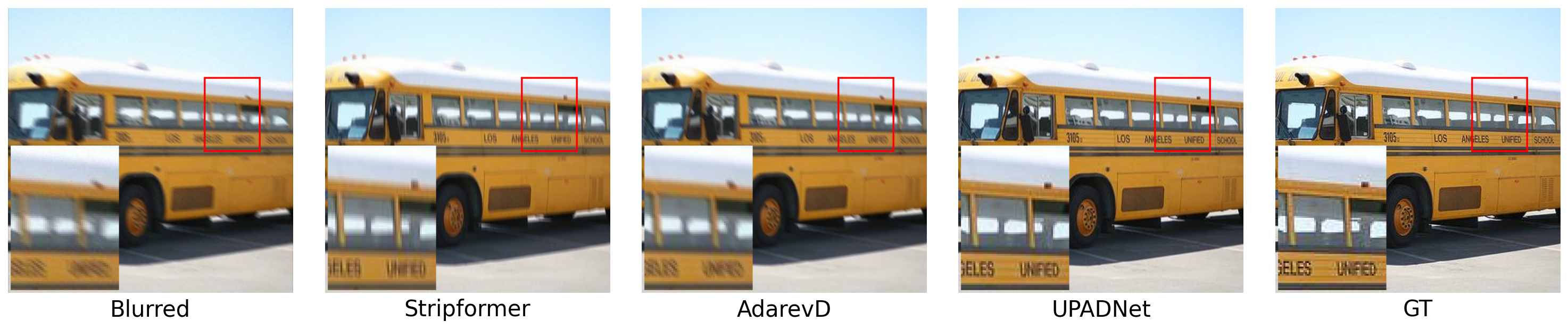}
  \caption{Qualitative results on the COCO dataset with 60\% of training data.
  }
  \label{fig:low-training}
\end{figure}

Table~\ref{tab:data_ratio} evaluates the robustness of different models under reduced training data ratios (100\%, 70\%, and 60\%) on the COCO dataset. As the amount of training data decreases, Stripformer and AdaRevD exhibit noticeable performance degradation, particularly in SSIM. In contrast, UPADNet maintains consistently high performance across all data regimes, achieving the best PSNR and SSIM at every training ratio. Notably, the performance drop from 100\% to 60\% training data is minimal for UPADNet, demonstrating strong data efficiency and improved generalization in low-data settings. Fig.~\ref{fig:low-training} presents qualitative comparisons under the 60\% training scenario. While competing methods suffer from residual blur and loss of fine structures, UPADNet better preserves edge sharpness and structural details, as evident in both Figs.~\ref{fig:COCO} and \ref{fig:low-training}. Because unrolled networks are derived from iterative algorithms---which typically require little or no training---UPADNet  demonstrates superior generalization ability.

\begin{table*}[t]
\caption{Comparison under reduced training data ratios (100\%, 70\%, 60\%) for COCO dataset.}
\label{tab:data_ratio}
\centering
\setlength{\tabcolsep}{6pt}
\renewcommand{\arraystretch}{1.15}
\begin{tabular}{l ccc ccc}
\toprule
\multirow{2}{*}[-3pt]{\textbf{Method}} & \multicolumn{3}{c}{\textbf{SSIM} $\uparrow$} 
& \multicolumn{3}{c}{\textbf{PSNR (dB)} $\uparrow$} \\
\cmidrule(lr){2-4} \cmidrule(lr){5-7}
& 100\% & 70\% & 60\%
& 100\% & 70\% & 60\% \\
\midrule

Stripformer \cite{tsai2022stripformer} 
& 0.921 & 0.825 & 0.805
& 31.84 & 28.34 & 27.21 \\

AdaRevD \cite{mao2024adarevd} 
& 0.925 & 0.896 & 0.865
& 32.10 & 31.43 & 29.51 \\

UPADNet 
& \textbf{0.932} & \textbf{0.929} & \textbf{0.926}
& \textbf{32.58} & \textbf{32.40} & \textbf{32.12} \\
\bottomrule
\end{tabular}
\end{table*}

\subsection{Ablation Study}
We conduct comprehensive ablation studies on the GoPro dataset to analyze the impact of (i) the number of UPADBlocks and (ii) the choice of activation function and learning rate.

\begin{table}[!t]
\caption{Ablation on the number of UPADBlocks in UPADNet.}
\label{tab:ablation_upadblocks}
\centering
\setlength{\tabcolsep}{8pt}
\renewcommand{\arraystretch}{1.15}
\begin{tabular}{c|ccccc}
\toprule
\# UPADBlocks & 24 & 36 & 48 & 60 & 72 \\
\midrule
PSNR $\uparrow$ & 28.41 & 29.68 & 32.89 & 33.02 & \textbf{34.08} \\
SSIM $\uparrow$ & 0.9512 & 0.9556 & 0.9581 & 0.9594 & \textbf{0.9601} \\
\bottomrule
\end{tabular}
\end{table}

Table~\ref{tab:ablation_upadblocks} evaluates the effects of progressively increasing the number of UPADBlocks from 24 to 72. As expected, increasing depth significantly improves restoration quality. The performance improves steadily from 28.41\,dB at 24 blocks to 34.08\,dB at 72 blocks, indicating that deeper unrolled structures better capture long-range dependencies and refine high-frequency details. Notably, the gains become moderate beyond 60 blocks, suggesting diminishing returns as the representational capacity saturates. This behavior is consistent with depth-scaling trends observed in modern restoration networks. Based on this trade-off between performance and computational cost, we select 72 blocks for the final model, as it provides the best quantitative performance.

Table~\ref{tab:ablation_act_lr_gopro} compares ReLU and GELU activations under two learning rates ($1\times10^{-3}$ and $1\times10^{-4}$). The results show that GELU consistently outperforms ReLU across both metrics. In particular, GELU with a learning rate of $1\times10^{-3}$ achieves the best performance (34.01\,dB PSNR and 0.951 SSIM), demonstrating improved optimization stability and smoother gradient propagation. Reducing the learning rate to $1\times10^{-4}$ slightly degrades performance, indicating slower convergence and reduced exploration of the parameter space. These findings suggest that smoother nonlinearities combined with moderately larger learning rates are beneficial for phase–amplitude reconstruction tasks.

\begin{table}[t]
\caption{Ablation on activation function and learning rate.}
\label{tab:ablation_act_lr_gopro}
\centering
\setlength{\tabcolsep}{7pt}
\renewcommand{\arraystretch}{1.15}
\begin{tabular}{c|cccc}
\toprule
\multirow{2}{*}[-3pt]{Metric} & \multicolumn{2}{c}{ReLU} & \multicolumn{2}{c}{GELU} \\
\cmidrule(lr){2-3}\cmidrule(lr){4-5}
& lr=$1{\times}10^{-3}$ & lr=$1{\times}10^{-4}$ & lr=$1{\times}10^{-3}$ & lr=$1{\times}10^{-4}$ \\
\midrule
PSNR $\uparrow$ & 29.52 & 31.18 & \textbf{34.01} & 33.10 \\
SSIM $\uparrow$ & 0.8638 & 0.9119 & \textbf{0.951} & 0.935 \\
\bottomrule
\end{tabular}
\end{table}

Overall, the ablation experiments demonstrate that both sufficient unrolled depth and appropriate activation design are critical for achieving state-of-the-art deblurring performance.

\section{Conclusion}


Our work develops UPADNet, a deep unrolled network architecture that jointly recovers the amplitude and phase components of a clean image from blurred and noisy observations. By decomposing the image and blur kernel into Fourier-domain amplitude and phase components, we derive LMMSE estimators for the amplitude and phase of the blurred image and integrated them into a regularized objective solved via alternating optimization and HQS. The resulting iterative scheme is then unfolded into a learnable network with dynamic, iteration-specific parameters.
\begin{sloppypar}
Extensive experiments on GoPro, RealBlur, and noise-corrupted COCO datasets demonstrate that UPADNet outperforms state-of-the-art alternatives. 
In particular, the explicit unrolled network for phase-component recovery---derived from an LMMSE estimation-inspired iterative formulation---is crucial for achieving performance gains in challenging real-world scenarios under varying blur type, different noise levels, and limited training data.
\end{sloppypar}

Overall, this work highlights the importance of structured frequency-domain modeling within deep unrolling frameworks and opens new directions for interpretable and physics-inspired learning in inverse imaging problems.

\section*{Acknowledgment}
V. Monga was supported by National Science Foundation (NSF) under Grant ECCS-2143557.

C. Lee's work was supported by the Institute of Information \& Communications Technology Planning \& Evaluation (IITP) under the Artificial Intelligence Convergence Innovation Human Resources Development (IITP-2026-RS-2023-00254592) grant funded by the Korea government (MSIT).

%
%
\newpage
\bibliographystyle{splncs04}
\bibliography{main}
\end{document}


\title{Leveraging Phase Information to Boost Unrolled Network Learning for Image Deblurring (Supplement)} 


\author{Samira Malek\inst{1}\orcidlink{0009-0005-2530-3846} \and
Haichuan Zhang\inst{1}\orcidlink{0009-0003-1678-6415} \and
Chul Lee\inst{2}\orcidlink{0000-0001-9329-7365} \and
Vishal Monga\inst{1}\orcidlink{0000-0002-5100-2263}}

\authorrunning{S.~Malek et al.}

\institute{Pennsylvania State University, University Park, PA 16802, USA
\email{\{sxm6547,hzz5333,vum4\}@psu.edu}\\
\and
Dongguk University, Seoul 04620, South Korea\\
\email{chullee@dongguk.edu}}

\maketitle

\appendix
\section*{Supplementary Material}
\setcounter{section}{0}
\renewcommand{\thesection}{S\arabic{section}}
\section{Proof of Theorem 1}
\begin{lemma}
\label{lem: LMMSE-2}
Let \( X \) and \( Y \) be two random variables with finite means and variances. 

\vspace{0.5em}
\textbf{Objective Function:}
\begin{equation}
 h(a, b) = \mathbb{E}[(X - aY - b)^2]   
\end{equation}

\vspace{0.5em}
\textbf{Then, \( h(a, b) \) is minimized when:}
\begin{equation}
    a^* = \frac{\mathrm{Cov}(X, Y)}{\mathrm{Var}(Y)}, \quad
b^* = \mathbb{E}[X] - a^* \mathbb{E}[Y]
\end{equation}
\begin{proof}
   See chapter 9 in \cite{pishro2014introduction}. 
\end{proof}

\end{lemma}

Now, we can prove theorem 1:\\
\begin{proof}
We aim to minimize the mean squared error between \( A_Z \) and its estimate \( \hat{A}_Z \) with respect to the weights \( \mathcal{W}_1 \) and \( \mathcal{W}_2 \):
\begin{equation}
    \min_{\mathcal{W}_1, \mathcal{W}_2} \; \mathbb{E}\left[ \|A_Z - \hat{A}_Z\|_F^2 \right]
    \quad \text{s.t.} \quad [\hat{A}_Z]_{i,j} = [\mathcal{W}_1]_{i,j} \circ [A_H \circ A_U]_{i,j} + [\mathcal{W}_2]_{i,j} .
\end{equation}

Since the squared error is separable across the index \( j \), the objective can be expressed as:
\begin{equation}
    \min_{\mathcal{W}_1, \mathcal{W}_2} \; \sum_i\sum_{j} 
    \mathbb{E} \left[ \big( [A_Z]_{i,j} - [\mathcal{W}_1]_{i,j} \, [A_H \circ A_U]_{i,j} - [\mathcal{W}_2]_{i,j} \big)^2 \right] .
\end{equation}

For each \( j \), this problem matches the form of the LMMSE problem in Lemma~\ref{lem: LMMSE-2}, with:
\[
X = [A_Z]_{i,j}, \quad Y = [A_H \circ A_U]_{i,j}, \quad a = [\mathcal{W}_1]_{i,j}, \quad b = [\mathcal{W}_2]_{i,j} .
\]

From Lemma~\ref{lem: LMMSE-2}, the minimizers for each \( j \) are:
\begin{equation}
    [\mathcal{W}_1^*]_{i,j} = \frac{\mathrm{Cov}([A_Z]_{i,j}, [A_H]_{i,j} \, [A_U]_{i,j})}
    {\mathrm{Var}([A_H]_{i,j} \, [A_U]_{i,j})}, 
    \quad
    [\mathcal{W}_2^*]_{i,j} = \mathbb{E}[[A_Z]_{i,j}] - [\mathcal{W}_1^*]_{i,j} \, \mathbb{E}[[A_H]_{i,j} \, [A_U]_{i,j}] .
\end{equation}

Therefore, applying the LMMSE solution to each index \( j \) yields the optimal weights \( \mathcal{W}_1^* \) and \( \mathcal{W}_2^* \) for minimizing the total error.
\end{proof}

\section{Proof of Theorem 2}
\begin{lemma}[LMMSE for three random variables]
\label{lem: LMMSE-3}
Let \( X \), \( Y_1 \), \( Y_2 \) be three random variables with finite means and variances. 

\begin{equation}
 h(a, b, c) = \mathbb{E}[(X - aY_1 - b Y_2 - c)^2]   
\end{equation}
Then, \( h(a, b, c) \) is minimized when:
\begin{equation}
\begin{aligned}
    a^* &= \frac{\sigma_2^2 \gamma_1 - \sigma_{12} \gamma_2}{\Delta}, \\
    b^* &= \frac{-\sigma_{12} \gamma_1 + \sigma_1^2 \gamma_2}{\Delta}, \\
    c^* &= \mu_X - a^* \mu_1 - b^* \mu_2
\end{aligned}
\end{equation}

\textit{where:}
\begin{align}
\sigma_1^2 &= \operatorname{Var}(Y_1), &
\sigma_2^2 &= \operatorname{Var}(Y_2), &
\sigma_{12} &= \operatorname{Cov}(Y_1, Y_2), \\
\gamma_1 &= \operatorname{Cov}(X, Y_1), &
\gamma_2 &= \operatorname{Cov}(X, Y_2), &
\Delta &= \sigma_1^2 \sigma_2^2 - \sigma_{12}^2, \\
\mu_X &= \mathbb{E}[X], &
\mu_1 &= \mathbb{E}[Y_1], &
\mu_2 &= \mathbb{E}[Y_2]
\end{align}
\end{lemma}

\begin{proof}
  We have:
    \begin{align}
h(a,b,c)
&=\mathbb{E}\!\left[(X-aY_1-bY_2-c)^2\right]\\
&=\mathbb{E}[X^2] + a^2\,\mathbb{E}[Y_1^2] + b^2\,\mathbb{E}[Y_2^2] + c^2
-2a\,\mathbb{E}[XY_1] -2b\,\mathbb{E}[XY_2] -2c\,\mathbb{E}[X]\\
&\quad\; +2ab\,\mathbb{E}[Y_1Y_2] + 2ac\,\mathbb{E}[Y_1] + 2bc\,\mathbb{E}[Y_2].
\end{align}
Thus, $h(a,b)$ is a quadratic function of $a$, $b$, and $c$. We take the derivatives with respect to $a$, $b$, and $c$ and set them to zero, so we obtain:

\begin{align}
\frac{\partial h}{\partial a}=0 &\;\Longrightarrow\; \mathbb{E}[Y_1^2]\,a+\mathbb{E}[Y_1Y_2]\,b+\mathbb{E}[Y_1]\,c=\mathbb{E}[XY_1],\\
\frac{\partial h}{\partial b}=0 &\;\Longrightarrow\; \mathbb{E}[Y_1Y_2]\,a+\mathbb{E}[Y_2^2]\,b+\mathbb{E}[Y_2]\,c=\mathbb{E}[XY_2],\\
\frac{\partial h}{\partial c}=0 &\;\Longrightarrow\; \mathbb{E}[Y_1]\,a+\mathbb{E}[Y_2]\,b+c=\mathbb{E}[X].\label{eq: c}
\end{align}

From the third equation in \eqref{eq: c}:
\begin{equation}
 c=\mu_X-a\mu_1-b\mu_2. 
 \label{eq: result-c}
\end{equation}

Substitute \eqref{eq: result-c} into the first two equations and regroup terms around covariances:

\begin{align}
\big(\mathbb{E}[Y_1^2]-\mu_1^2\big)a+\big(\mathbb{E}[Y_1Y_2]-\mu_1\mu_2\big)b
&=\mathbb{E}[XY_1]-\mu_X\mu_1,\\
\big(\mathbb{E}[Y_1Y_2]-\mu_1\mu_2\big)a+\big(\mathbb{E}[Y_2^2]-\mu_2^2\big)b
&=\mathbb{E}[XY_2]-\mu_X\mu_2.
\end{align}

That is,
\begin{equation}
  \begin{pmatrix}
\sigma_1^2 & \sigma_{12}\\
\sigma_{12} & \sigma_2^2
\end{pmatrix}
\begin{pmatrix} a\\ b\end{pmatrix}
=
\begin{pmatrix} \gamma_1\\ \gamma_2\end{pmatrix},  
\end{equation}

Assuming $Y_1$ and $Y_2$ are not collinear (so $\Delta=\sigma_1^2\sigma_2^2-\sigma_{12}^2>0$), we get:
\begin{equation}
 a^*=\frac{\sigma_2^2\gamma_1-\sigma_{12}\gamma_2}{\Delta},\qquad
b^*=\frac{-\sigma_{12}\gamma_1+\sigma_1^2\gamma_2}{\Delta}.    
\end{equation}

Then from \eqref{eq: result-c},
\begin{equation}
  c^*=\mu_X-a^*\mu_1-b^*\mu_2.  
\end{equation}

This $(a^*,b^*,c^*)$ uniquely minimizes $h$.
\end{proof}
\begin{proof}
We aim to minimize the mean squared error between \( \theta_Z \) and its estimate \( \hat{\theta}_Z \) with respect to the weights \( \mathcal{W}_3, \mathcal{W}_4, \mathcal{W}_5 \):
\begin{equation}
    \min_{\mathcal{W}_3, \mathcal{W}_4, \mathcal{W}_5} \; \mathbb{E}\left[ \|\theta_Z - \hat{\theta}_Z\|_F^2 \right],
    \quad \text{s.t.} \quad 
    [\hat{\theta}_Z]_{i,j} = [\mathcal{W}_3]_{i,j} \, [\theta_H]_{i,j} + [\mathcal{W}_4]_{i,j} \, [\theta_U]_{i,j} + [\mathcal{W}_5]_{i,j} .
\end{equation}

Since the squared error is separable across the index \( j \), the objective can be written as:
\begin{equation}
    \min_{\mathcal{W}_3, \mathcal{W}_4, \mathcal{W}_5} \; \sum_i\sum_{j}
    \mathbb{E} \left[ \big( [\theta_Z]_{i,j} - [\mathcal{W}_3]_{i,j} [\theta_H]_{i,j} - [\mathcal{W}_4]_{i,j} [\theta_U]_{i,j} - [\mathcal{W}_5]_{i,j} \big)^2 \right] .
\end{equation}

For each \( j \), this problem matches the form of the LMMSE problem in Lemma~\ref{lem: LMMSE-3}, with:
\[
X = [\theta_Z]_{i,j}, \quad 
Y_1 = [\theta_H]_{i,j}, \quad 
Y_2 = [\theta_U]_{i,j}, \quad 
a = [\mathcal{W}_3]_{i,j}, \quad 
b = [\mathcal{W}_4]_{i,j}, \quad 
c = [\mathcal{W}_5]_{i,j} .
\]

From Lemma~\ref{lem: LMMSE-3}, the minimizers for each \( j \) are:
\begin{align}
    [\mathcal{W}_3^*]_{i,j} &= \frac{\sigma_{2,j}^2 \gamma_{1,j} - \sigma_{12,j} \gamma_{2,j}}{\Delta_{i,j}}, \\
    [\mathcal{W}_4^*]_{i,j} &= \frac{-\sigma_{12,j} \gamma_{1,j} + \sigma_{1,j}^2 \gamma_{2,j}}{\Delta_{i,j}}, \\
    [\mathcal{W}_5^*]_{i,j} &= \mu_{\theta_Z,j} - [\mathcal{W}_3^*]_{i,j} \, \mu_{\theta_H,j} - [\mathcal{W}_4^*]_{i,j} \, \mu_{\theta_U,j},
\end{align}
where:
\begin{align}
\sigma_{1,j}^2 &= \operatorname{Var}([\theta_H]_{i,j}), &
\sigma_{2,j}^2 &= \operatorname{Var}([\theta_U]_{i,j}), &
\sigma_{12,j} &= \operatorname{Cov}([\theta_H]_{i,j}, [\theta_U]_{i,j}), \\
\gamma_{1,j} &= \operatorname{Cov}([\theta_Z]_{i,j}, [\theta_H]_{i,j}), &
\gamma_{2,j} &= \operatorname{Cov}([\theta_Z]_{i,j}, [\theta_U]_{i,j}), &
\Delta_{i,j} &= \sigma_{1,j}^2 \sigma_{2,j}^2 - \sigma_{12,j}^2, \\
\mu_{\theta_Z,j} &= \mathbb{E}[[\theta_Z]_{i,j}], &
\mu_{\theta_H,j} &= \mathbb{E}[[\theta_H]_{i,j}], &
\mu_{\theta_U,j} &= \mathbb{E}[[\theta_U]_{i,j}] .
\end{align}

Thus, applying the three-variable LMMSE solution to each index \( j \) yields the optimal weight vectors \( \mathcal{W}_3^*, \mathcal{W}_4^*, \mathcal{W}_5^* \) for minimizing the total error.
\end{proof}

\section{Extension to RGB Images}
\label{sec:rgb_extension}

For an RGB image, we denote the blurred observation by
\[
\mathbf{z}=\{z^{(r)},z^{(g)},z^{(b)}\},
\]
where $z^{(r)}$, $z^{(g)}$, and $z^{(b)}$ represent the red, green, and blue channels, respectively. Let the corresponding latent sharp image be
\[
\mathbf{u}=\{u^{(r)},u^{(g)},u^{(b)}\}.
\]
Assuming a physically consistent blur formation model, the blur kernel is shared across all channels, while the latent image is channel-dependent:
\begin{equation}
z^{(c)} = h * u^{(c)} + n^{(c)}, \qquad c\in\{r,g,b\},
\end{equation}
where $h$ is the common blur kernel and $n^{(c)}$ denotes the additive noise in channel $c$.

After Fourier-domain phase-amplitude decomposition, the amplitude and phase variables for each channel are denoted by
\begin{equation}
 A_Z^{(c)},\quad A_U^{(c)},\quad \theta_Z^{(c)},\quad \theta_U^{(c)},   
\end{equation}
while the blur kernel is represented by its amplitude and phase components
\begin{equation}
A_H,\quad \theta_H.   
\end{equation}
Using the same amplitude-phase parameterization as in the grayscale case, the RGB objective can be written as
\begin{align}
\label{eq:rgb_objective}
\min_{\substack{A_H,\theta_H,\\ \{A_U^{(c)},\theta_U^{(c)}\}_{c\in\{r,g,b\}}}}
\quad
&\sum_{c\in\{r,g,b\}}
\Big(
\|A_Z^{(c)}-(\mathcal{W}_1^{(c)}\circ A_H \circ A_U^{(c)}+\mathcal{W}_2^{(c)})\|_F^2
\Big)
+ \alpha \|A_H\|_F^2
\nonumber\\& + \beta \sum_{c\in\{r,g,b\}} \|M\circ A_U^{(c)}\|_{1,1}
\nonumber\\
&\quad +
\sum_{c\in\{r,g,b\}}
\Big(
\|\theta_Z^{(c)}-(\mathcal{W}_3^{(c)}\circ \theta_H + \mathcal{W}_4^{(c)}\circ \theta_U^{(c)}+\mathcal{W}_5^{(c)})\|_F^2
\Big)
\nonumber\\&+ \gamma \|\theta_H\|_F^2
+ \mu \sum_{c\in\{r,g,b\}} \|\theta_U^{(c)}\|_{1,1} .
\end{align}
Here, $\alpha,\beta,\gamma,\mu>0$ are regularization parameters, and $M$ is a frequency-domain weighting mask used to promote structured sparsity in the latent amplitude.

The main difficulty in \eqref{eq:rgb_objective} arises from the $\ell_1$ penalties on $A_U^{(c)}$ and $\theta_U^{(c)}$. To decouple the nonsmooth terms from the quadratic fidelity terms, we use HQS method and introduce auxiliary variables:
\begin{equation}
  S^{(c)} \approx A_U^{(c)}, \qquad P^{(c)} \approx \theta_U^{(c)}, \qquad c\in\{r,g,b\}.  
\end{equation}

This leads to the equivalent alternating minimization formulation:
\begin{align}
\label{eq:rgb_aux_objective}
\min_{\substack{A_H,\theta_H,\\ \{A_U^{(c)},\theta_U^{(c)},S^{(c)},P^{(c)}\}_{c\in\{r,g,b\}}}}
\quad
&\sum_{c\in\{r,g,b\}}
\|A_Z^{(c)}-(\mathcal{W}_1^{(c)}\circ A_H \circ A_U^{(c)}+\mathcal{W}_2^{(c)})\|_F^2
+ \alpha \|A_H\|_F^2
\nonumber\\
&\quad
+ \beta \sum_{c\in\{r,g,b\}} \|A_U^{(c)}-S^{(c)}\|_F^2
+ \zeta \sum_{c\in\{r,g,b\}} \|M\circ S^{(c)}\|_{1,1}
\nonumber\\
&\quad
+ \sum_{c\in\{r,g,b\}}
\|\theta_Z^{(c)}-(\mathcal{W}_3^{(c)}\circ \theta_H + \mathcal{W}_4^{(c)}\circ \theta_U^{(c)}+\mathcal{W}_5^{(c)})\|_F^2
\nonumber\\
&\quad
+ \gamma \|\theta_H\|_F^2+ \mu \sum_{c\in\{r,g,b\}} \|\theta_U^{(c)}-P^{(c)}\|_F^2
\nonumber\\&+ \xi \sum_{c\in\{r,g,b\}} \|P^{(c)}\|_{1,1},
\end{align}
where $\zeta>0$ and $\xi>0$ control the sparsity level of the auxiliary variables.

With all other variables fixed, \eqref{eq:rgb_aux_objective} can be minimized by alternating updates over six varibales: the amplitude of kernel$A_H$, the per-channel latent amplitudes $A_U^{(c)}$, the amplitude auxiliary variables $S^{(c)}$, the phase of kernel $\theta_H$, the per-channel latent phases $\theta_U^{(c)}$, and the phase auxiliary variables $P^{(c)}$.
\begin{itemize}
    \item \textbf{The amplitude of kernel update:}
Fixing $\{A_U^{(c,t)}\}_{c}$, the update of the kernel amplitude is
\begin{equation}
\label{eq:rgb_AH_subproblem}
A_H^{(t+1)}
=
\arg\min_{A_H}
\sum_{c\in\{r,g,b\}}
\|A_Z^{(c)}-\mathcal{W}_1^{(c)}\circ A_H \circ A_U^{(c,t)}-\mathcal{W}_2^{(c)}\|_F^2
+\alpha \|A_H\|_F^2.
\end{equation}
Since this is a quadratic problem in $A_H$, its minimizer is obtained in closed form:
\begin{equation}
\label{eq:rgb_AH_closed}
A_H^{(t+1)}=
\frac{
\sum\limits_{c\in\{r,g,b\}}
(\mathcal{W}_1^{(c)}\circ A_U^{(c,t)})
\circ
(A_Z^{(c)}-\mathcal{W}_2^{(c)})
}{
\sum\limits_{c\in\{r,g,b\}}
(\mathcal{W}_1^{(c)}\circ A_U^{(c,t)})^2
+\alpha }.
\end{equation}
\item \textbf{Amplitude auxiliary-variable update:}
For each channel $c$, fixing $A_U^{(c,t)}$, we solve
\begin{equation}
\label{eq:rgb_S_subproblem}
S^{(c,t+1)}
=
\arg\min_{S^{(c)}}
\zeta \|M\circ S^{(c)}\|_{1,1}
+
\beta \|A_U^{(c,t)}-S^{(c)}\|_F^2.
\end{equation}
This is a standard proximal $\ell_1$ problem, whose solution is given by element-wise soft-thresholding:
\begin{equation}
\label{eq:rgb_S_closed}
S^{(c,t+1)}
=
\operatorname{soft}
\left(
A_U^{(c,t)},
\frac{\zeta |M|}{2\beta}
\right),
\end{equation}
where
\begin{equation}
\operatorname{soft}(x,\tau)=\operatorname{sign}(x)\max(|x|-\tau,0)   
\end{equation}
is applied element-wise.
\item \textbf{ Per-channel latent image amplitude update:}
Given $A_H^{(t+1)}$ and $S^{(c,t+1)}$, each channel amplitude is updated by
\begin{equation}
\label{eq:rgb_AU_subproblem}
A_U^{(c,t+1)}
=
\arg\min_{A_U^{(c)}}
\|A_Z^{(c)}-\mathcal{W}_1^{(c)}\circ A_H^{(t+1)} \circ A_U^{(c)}-\mathcal{W}_2^{(c)}\|_F^2
+
\beta \|A_U^{(c)}-S^{(c,t+1)}\|_F^2.
\end{equation}
Again, this is a quadratic problem and admits the following closed-form solution:
\begin{equation}
\label{eq:rgb_AU_closed}
A_U^{(c,t+1)}
=
\frac{
(\mathcal{W}_1^{(c)}\circ A_H^{(t+1)})
\circ
(A_Z^{(c)}-\mathcal{W}_2^{(c)})
+
\beta S^{(c,t+1)}
}{
(\mathcal{W}_1^{(c)}\circ A_H^{(t+1)})^2+\beta }.
\end{equation}
\item \textbf{The phase of kernel update:}
Fixing $\{\theta_U^{(c,t)}\}_{c}$, the phase of kernel is updated by
\begin{equation}
\label{eq:rgb_thetaH_subproblem}
\theta_H^{(t+1)}
=
\arg\min_{\theta_H}
\sum_{c\in\{r,g,b\}}
\|\theta_Z^{(c)}-\mathcal{W}_3^{(c)}\circ \theta_H-\mathcal{W}_4^{(c)}\circ \theta_U^{(c,t)}-\mathcal{W}_5^{(c)}\|_F^2
+\gamma \|\theta_H\|_F^2.
\end{equation}
This subproblem is also quadratic, yielding
\begin{equation}
\label{eq:rgb_thetaH_closed}
\theta_H^{(t+1)}=
\frac{
\sum\limits_{c\in\{r,g,b\}}
\mathcal{W}_3^{(c)} \circ
\left(
\theta_Z^{(c)}-(\mathcal{W}_4^{(c)}\circ\theta_U^{(c,t)}+\mathcal{W}_5^{(c)})
\right)
}{
\sum\limits_{c\in\{r,g,b\}}
(\mathcal{W}_3^{(c)})^2+\gamma }.
\end{equation}
\item \textbf{Phase auxiliary-variable update:}
For each channel $c$, fixing $\theta_U^{(c,t)}$, we solve
\begin{equation}
\label{eq:rgb_P_subproblem}
P^{(c,t+1)}
=
\arg\min_{P^{(c)}}
\xi \|P^{(c)}\|_{1,1}
+
\mu \|\theta_U^{(c,t)}-P^{(c)}\|_F^2.
\end{equation}
Its solution is the element-wise soft-thresholding operator
\begin{equation}
\label{eq:rgb_P_closed}
P^{(c,t+1)}
=
\operatorname{soft}
\left(
\theta_U^{(c,t)},
\frac{\xi}{2\mu}
\right).
\end{equation}
\item \textbf{Per-channel phase update:}
Given $\theta_H^{(t+1)}$ and $P^{(c,t+1)}$, each latent phase is updated via
\begin{equation}
\label{eq:rgb_thetaU_subproblem}
\theta_U^{(c,t+1)}
=
\arg\min_{\theta_U^{(c)}}
\|\theta_Z^{(c)}-\mathcal{W}_3^{(c)}\circ \theta_H^{(t+1)}-\mathcal{W}_4^{(c)}\circ \theta_U^{(c)}-\mathcal{W}_5^{(c)}\|_F^2
+
\mu \|\theta_U^{(c)}-P^{(c,t+1)}\|_F^2.
\end{equation}
This subproblem is quadratic and has the closed-form solution
\begin{equation}
\label{eq:rgb_thetaU_closed}
\theta_U^{(c,t+1)}
=
\frac{
\mathcal{W}_4^{(c)} \circ
\left(
\theta_Z^{(c)}-\mathcal{W}_3^{(c)}\circ\theta_H^{(t+1)}-\mathcal{W}_5^{(c)}
\right)
+
\mu P^{(c,t+1)}
}{
(\mathcal{W}_4^{(c)})^2+\mu }.
\end{equation}
\end{itemize}
Therefore, each iteration of the alternating minimization algorithm consists only of closed-form updates, making the overall solver efficient and well suited for unrolling into the proposed UPADNet architecture. This kernel formulation also improves estimation stability, since the blur kernel is constrained using complementary information from all RGB channels. Consequently, \cref{alg:simplified-opad-rgb} summarizes the complete RGB deblurring procedure and provides the update rules used in the network design.

\begin{algorithm}[t]
\caption{Phase and Amplitude Decomposition Image Deblurring Algorithm (RGB Image)}
\label{alg:simplified-opad-rgb}
\begin{algorithmic}[1]
\Require $\{A_Z^{(c)}, \theta_Z^{(c)}\}_{c\in\{r,g,b\}}$, $\{\mathcal{W}_i^{(c)}\}_{i=1}^5$ (optional per-channel), filter $M$, regularization weights $\alpha,\beta,\gamma,\mu,\xi,\zeta$
\State \textbf{Initialize:} $\{A_U^{(c,0)},\theta_U^{(c,0)}\}_{c\in\{r,g,b\}}$, and shared kernel variables $A_H^{(0)},\theta_H^{(0)}$
\For{$t=0$ to $T$}

\Statex \textbf{Amplitude updates (shared kernel, per-channel image):}

\State \hspace{1em}$
A_H^{(t+1)} =
\frac{
\sum\limits_{c\in\{r,g,b\}}
\left(\mathcal{W}_1^{(c)} \circ A_U^{(c,t)}\right)\circ\left(A_Z^{(c)}-\mathcal{W}_2^{(c)}\right)
}{
\sum\limits_{c\in\{r,g,b\}}
\left(\mathcal{W}_1^{(c)} \circ A_U^{(c,t)}\right)^2
+\alpha
}$

\For{$c\in\{r,g,b\}$}
    \State \hspace{1em}$
    S^{(c,t+1)} =
    \operatorname{soft}\!\left(
      A_U^{(c,t)}, \frac{\zeta \circ |M|}{2\beta}
    \right)$

    \State \hspace{1em}$
    A_U^{(c,t+1)} =
    \frac{
      \left(\mathcal{W}_1^{(c)} \circ A_H^{(t+1)}\right)\circ
      \left(A_Z^{(c)}-\mathcal{W}_2^{(c)}+\beta S^{(c,t+1)}\right)
    }{
      \left(\mathcal{W}_1^{(c)} \circ A_H^{(t+1)}\right)^2+\beta
    }$
\EndFor

\Statex \textbf{Phase updates (shared kernel, per-channel image):}

\State \hspace{1em}$
\theta_H^{(t+1)} =
\frac{
\sum\limits_{c\in\{r,g,b\}}
\mathcal{W}_3^{(c)} \circ
\left(\theta_Z^{(c)}-\left(\mathcal{W}_4^{(c)}\circ\theta_U^{(c,t)}+\mathcal{W}_5^{(c)}\right)\right)
}{
\sum\limits_{c\in\{r,g,b\}}
\left(\mathcal{W}_3^{(c)}\right)^2
+\gamma
}$

\For{$c\in\{r,g,b\}$}
    \State \hspace{1em}$
    P^{(c,t+1)} =
    \operatorname{soft}\!\left(
      \theta_U^{(c,t)}, \frac{\xi}{2\mu}
    \right)$

    \State \hspace{1em}$
    \theta_U^{(c,t+1)} =
    \frac{
      \mathcal{W}_4^{(c)} \circ
      \left(\theta_Z^{(c)}-\left(\mathcal{W}_3^{(c)}\circ\theta_H^{(t+1)}+\mathcal{W}_5^{(c)}\right)+\mu P^{(c,t+1)}\right)
    }{
      \left(\mathcal{W}_4^{(c)}\right)^2+\mu
    }$
\EndFor

\EndFor
\Ensure $\{A_U^{(c,T+1)},\theta_U^{(c,T+1)}\}_{c\in\{r,g,b\}}$ and shared $A_H^{(T+1)},\theta_H^{(T+1)}$
\end{algorithmic}
\end{algorithm}

\bibliographystyle{splncs04}
\bibliography{main}